\documentclass[letterpaper, 10 pt, conference]{ieeeconf}

\IEEEoverridecommandlockouts                              %

\let\labelindent\relax

\usepackage{graphicx} %
\usepackage{subcaption}
\usepackage{enumerate}
\usepackage{enumitem}
\usepackage{microtype}
\usepackage{algorithm}
\usepackage{xcolor}
\usepackage{cite}
\usepackage{flushend}
\usepackage{xcolor}
\usepackage[breaklinks,colorlinks]{hyperref}
\usepackage[hang,flushmargin,symbol]{footmisc}
\usepackage{amsmath}
\usepackage{amssymb}
\usepackage{tensor}

\DeclareMathAlphabet{\mathcal}{OMS}{cmsy}{m}{n}

\def\R{\ensuremath{\mathbb{R}}}

\def\A{\ensuremath{\mathcal{A}}}
\def\D{\ensuremath{\mathcal{D}}}

\def\M{\ensuremath{\mathcal{M}}}

\def\O{\ensuremath{\mathcal{O}}}

\def\S{\ensuremath{\mathcal{S}}}
\def\T{\ensuremath{\mathcal{T}}}

\newcommand{\mat}[1]{\ensuremath{\mathbf{#1}}}

\renewcommand{\det}[1]{\left|#1\right|}
\newcommand{\cardinality}[1]{\left|#1\right|}

\newcommand{\prob}[1]{\ensuremath{p\left(#1\right)}}
\newcommand{\probc}[2]{\ensuremath{\prob{#1 \;\middle\vert\; #2}}}
\newcommand{\probdist}[2]{\ensuremath{p_{#1}\left(#2\right)}}

\newcommand{\set}[1]{\ensuremath{\left\{#1\right\}}}
\newcommand{\fset}[2]{\ensuremath{\set{#1 \;\middle\vert\; #2}}}

\newcommand{\tf}[3]{\tensor[^{#1}]{\mat{#2}}{_{#3}}}

\newcommand{\tuple}[1]{\left\langle #1\right\rangle}

\DeclareMathOperator*{\argmin}{arg\,min}

\usepackage{xspace}

\newcommand{\ie}{\mbox{i.\,e.}\xspace}

\newcommand{\etal}{\emph{et al.}\xspace}
   
\renewcommand{\[}{\begin{equation}}
\renewcommand{\]}{\end{equation}}

\renewcommand{\baselinestretch}{0.99}

\usepackage[capitalize]{cleveref}

\crefname{figure}{Fig.}{Figs.}
\Crefname{figure}{Figure}{Figures}
\crefname{section}{Sec.}{Secs.}
\Crefname{section}{Section}{Sections}
\Crefname{table}{Table}{Tables}
\crefname{table}{Tab.}{Tabs.}
\crefname{algorithm}{Algo.}{Algos.}
\Crefname{algorithm}{Algorithm}{Algorithms}
\crefname{appendix}{Sec.}{Secs.}
\Crefname{appendix}{Section}{Sections}

\title{SparTa: Sparse Graphical Task Models from a Handful of Demonstrations}
\author{Adrian Röfer$^{1}$, Nick Heppert$^{1,2}$, and Abhinav Valada$^{1}$
\thanks{\mbox{$^1$}University of Freiburg, Germany, 
        \mbox{$^2$Zuse} School ELIZA}
\thanks{\noindent This work was funded by the Carl Zeiss Foundation with the ReScaLe
project and the BrainLinks-BrainTools center of the University of Freiburg.
Nick Heppert is supported by the Konrad Zuse School of Excellence in
Learning and Intelligent Systems (ELIZA) through the DAAD programme
Konrad Zuse Schools of Excellence in Artificial Intelligence, sponsored by
the Federal Ministry of Education and Research.}
\thanks{The authors would like to thank Karla Štěpánová for taking the time to discuss the method and giving feedback on the manuscript. We would also like to thank Maurice Funk for his consultation on $k$-AP matching.}
}

\makeatletter
\g@addto@macro{\endtabular}{\rowfont{}}%
\makeatother
\newcommand{\rowfonttype}{}%
\newcommand{\rowfont}[1]{%
\gdef\rowfonttype{#1}#1\ignorespaces
}

\DeclareMathOperator{\eAdd}{\textsc{Add}}
\DeclareMathOperator{\eMerge}{\textsc{Merge}}
\DeclareMathOperator{\eSplit}{\textsc{Split}}

\newif\ifmix
\mixtrue %

\begin{document}
\bstctlcite{IEEEexample:BSTcontrol} %
\maketitle
\pagestyle{empty}

\begin{abstract}
   Learning long-horizon manipulation tasks efficiently is a central challenge in robot learning from demonstration. Unlike recent endeavors that focus on directly learning the task in the action domain, we focus on inferring \emph{what} the robot should achieve in the task, rather than \emph{how} to do so.
   To this end, we represent evolving scene states using a series of graphical object relationships. We propose a demonstration segmentation and pooling approach that extracts a series of manipulation graphs and estimates distributions over object states across task phases. In contrast to prior graph-based methods that capture only partial interactions or short temporal windows, our approach captures complete object interactions spanning from the onset of control to the end of the manipulation. To improve robustness when learning from multiple demonstrations, we additionally perform object matching using pre-trained visual features. In extensive experiments, we evaluate our method's demonstration segmentation accuracy and the utility of learning from multiple demonstrations for finding a desired minimal task model. Finally, we deploy the fitted models both in simulation and on a real robot, demonstrating that the resulting task representations support reliable execution across environments.
\end{abstract}

\section{Introduction}
\label{section:intro}

Learning long-horizon tasks such as assembling a shelf, changing a tire, or preparing a meal, motion by motion, is a daunting prospect. If learning focuses exclusively on \emph{how} a task is performed, it becomes easy to lose sight of \emph{what} the task is trying to accomplish: its goals and sub-objectives. 
Arguably, much of the focus in robotic manipulation learning from demonstration has been on \emph{how} rather than \emph{what}. Behavior cloning methods map raw state observations to momentary actions~\cite{chi2024diffusion,chisari2024learning,honerkamp2025whole}. They can excel on short-horizon and even dexterous tasks, but their high-dimensional input-output mappings typically require tens to hundreds of demonstrations to generalize reliably.

A complementary perspective views robotic manipulation primarily as acting on objects and the external world rather than reproducing motions themselves, a view formalized early in \emph{Object Action Complexes} (OAC)~\cite{kruger2011object}. This object-centric perspective has also served as a prior for enhancing modern behavior cloning methods~\cite{rana2024affordance,von2024art,heppert2024ditto,montero2024learning}. Such methods segment demonstrations into motion segments, either by using gripper state changes or low-velocity phases, and establish object-aligned coordinate frames in which the demonstrated motion, \ie \emph{how}, is represented. With reliable frame re-detection, these approaches can reduce sample complexity and improve robustness to scene variation. However, even after executing the learned motions, they typically cannot determine whether the intended objective state has been achieved, because the objective, the \emph{what}, is not explicitly modeled. %

\begin{figure}
    \centering
    \includegraphics[width=0.9\linewidth]{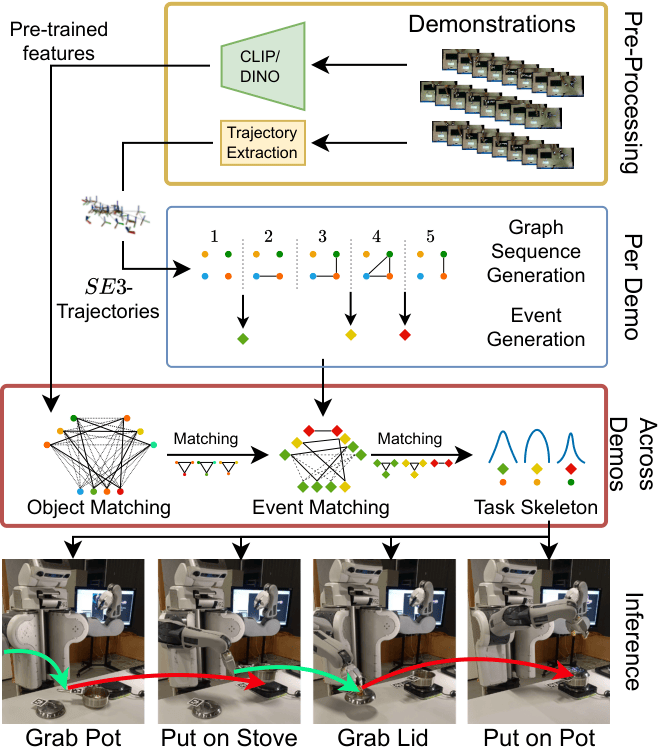}
    \caption{Our approach extracts sparse task skeletons from demonstrations. Using object trajectories, it builds a series of manipulation graphs and generates events over graph changes. Using pre-trained features, objects are matched across demonstrations, the events of the same objects are grouped, and a task skeleton is extracted. At inference, these events are interpreted as grasping and placement actions, and the target poses are inferred from the extracted distributions.
    }
    \label{fig:approach}
    \vspace{-3mm}
\end{figure}

In this work, we therefore focus on learning \emph{what} a demonstration is trying to achieve. Concretely, we aim to learn that the toast goes into the toaster, then onto a plate, which is placed on a tray with a coffee cup. In task and motion planning (TAMP) terms, our goal is to automatically extract probabilistic post-conditions from a set of task demonstrations~\cite{garrett2021integrated}, yielding an explicit task model that supports planning at the level of object relations.

The approach we present in this work uses a graphical representation of scene states formulated as objects with changing relationships, a common abstraction for goal learning~\cite{aksoy2011learning,zhu2024vision,merlo2025exploiting,wang2025oneshot}. Early formulations include rich semantic relationships~\cite{aksoy2011learning}, while subsequent work has increasingly simplified relations to binary physical contact~\cite{aein2019library,zhu2024vision,wang2025oneshot}.
Most recently, Merlo~\etal~\cite{merlo2025exploiting} proposed building a contact graph whose edges form on the basis of manipulation as evidenced by the \emph{Mutual Information} (MI) of different object trajectories. This interpretation likens the object graphs more to kinematic graphs rather than abstract semantic graphs.

Adding to this development, we contribute an approach with three key advances:
\begin{itemize}
    \item \textit{Persistent manipulation edges} harden the concept of~\cite{merlo2025exploiting} into manipulation graphs, which capture when control is first established and lasts until the manipulation is concluded, strengthening the idea of kinematic graphs. %
    \item \textit{Feature-based object matching} enables learning from multiple demonstrations, which allows us to capture variance in the demonstrations and retrieve relevant background objects without a prior. Different from~\cite{aksoy2011learning}, our approach does not assume known object associations for this process, but matches objects based on the similarity of pre-trained feature vectors, which can be obtained easily and reliably~\cite{radford2021learning,oquab2023dinov2}. 
    \item \textit{Learning from a handful of demonstrations} is enabled by simplifying the matching process to only consider transitions of objects between independent kinematic subgraphs. %
    These changes are matched across demonstrations to find the tightest distribution of pre- and post-conditions when this change occurs. In the end, our full approach yields a series of \emph{activations} and \emph{deactivations} of objects and their probable poses at these moments.
\end{itemize}
In this paper, we provide detailed insights into the strengths and weaknesses of our proposed method. To this end, we evaluate on the recent dataset of Merlo~\etal~\cite{merlo2025exploiting} and show success when using demonstrations from a robotic simulation benchmark primarily designed to investigate the \emph{how} rather than the \emph{what}. We study the quality of the segmentation and the benefits of pooling data from multiple demonstrations. Finally, we deploy our extracted models on a real robot.
Upon acceptance, we will release the code and data, including the newly created labels for this study at \url{https://sparta.cs.uni-freiburg.de}, to facilitate future research.

\section{Related Work}

Understanding manipulation tasks not primarily as motions but as modifications to the relationships between objects has yielded a number of approaches that model these relationships as edges in graphs over time. 

Aksoy~\etal~\cite{aksoy2011learning} presented the seminal concept of \emph{Semantic Event Chains} (SEC), which captures changing semantic relationships between objects over time. The relationships are image-space metrics such as \emph{touching}, \emph{overlapping}, \emph{not-touching}. They demonstrated that this representation, despite its sparsity, can be used to classify tasks or objects when additional information is provided. In a long line of works, the utility of SECs has been demonstrated for plan representation, as well as more detailed action understanding~\cite{aksoy2011execution,aksoy2015modelfree,aein2019library,ziaeetabar2024hierarchical,erdogan2025real}. A challenge of SECs is that \emph{contact} is the fundamental basis for object relationships, which is difficult to observe under occlusions or noisy image data. In addition, the association of demonstrations is performed by matching complete graph topologies. Given that subtasks are executed independently in the same demonstration, this increases the demand for training data. In contrast, our approach decouples kinematically independent subgraphs and associates them independently.

For a long time, SECs seem to have been the only approach to abstracting tasks as graphical object relationships. Recently, a few more approaches have revisited this concept~\cite{wang2025oneshot,zhu2024vision,merlo2025exploiting}. Wang~\etal~\cite{wang2025oneshot} introduce Parameterized Symbolic Abstraction Graphs (PSAGs), which capture temporal contact relationships between objects, similar to SECs. They use a multi-camera setup and modern instance segmentation and object pose tracking techniques to perform one-shot imitation of demonstrations. Zhu~\etal~\cite{zhu2024vision} present a similar method with less technical overhead. Their approach, ORION, extracts similar graphs from a single video, similar to PSAG. However, where PSAG uses contact to filter background objects, ORION requires language annotation. Given a selected set of objects, it performs velocity-based segmentation of manipulation segments and extracts reference frames for these segments.
Merlo \etal~\cite{merlo2025exploiting} criticize velocity-based segmentation and propose Mutual Information (MI) of object trajectories instead to determine if they are being manipulated or not. Using this concept, they propose a one-shot imitation approach which they demonstrate across a variety of tasks, and expand in a second work \cite{merlo2025information} to bi-manual manipulation tasks. Their approach detects the beginnings and ends of manipulations and clones the relative poses of objects at these moments, which can then be replayed with behavior trees and pre-defined motion primitives. 

In this work, we build on the insight of Merlo~\etal~\cite{merlo2025exploiting} on the utility of mutual information, but we reduce the number of modeling assumptions in the segmentation framework, such as distance thresholds for object interactions by fitting interaction models dynamically from the training data. In addition, unlike the aforementioned recent methods, we restore the ability to learn across multiple demonstrations. This enables us to use simple entropy to discriminate against distractor objects. Different from~\cite{aksoy2011learning,ziaeetabar2024hierarchical,erdogan2025real}, our approach does not require distinct object labels but instead uses a combination of role and pre-trained features to associate objects across different demonstrations.

A few works have also considered the problem of learning task scheduling constraints from demonstrations~\cite{shah2018bayesian,chou2022learning,dreher2024learning}. While interesting and relevant for dynamically scheduling different subtasks at inference time, we do not consider it in this work. Our proposed approach extracts strictly linear models, even if not all demonstrations exhibit the same sequence. Separating the different subtasks and their dependencies from these sequences is left for future work. 

In this work, we do not integrate language as a modality for the time being. Nonetheless, state-of-the-art LLMs possess substantial prior knowledge of everyday tasks and can be used to identify regions of task-relevance~\cite{zhao2025anyplace,fang2025kalm}. More prominently, language-conditioning has emerged as a popular method for encoding semantic goals in the Vision Language Action (VLA) family of behavior cloning approaches~\cite{octo2023,kim2024openvla,black2410pi0}. While these approaches are very good at dexterous manipulation tasks, it has recently been called into doubt whether the language conditioning had the intended effect~\cite{fei2025libero}.
In terms of model structure, these approaches are far removed from the one we present here. In the future, we would like to introduce language to steer our approach but not for now.

\section{Problem Definition}
\label{sec:problem}

We seek to learn a sparse model of a task that captures the manipulation relationships of objects over time and records the relative state of objects when their relationships change. A task $\T$ is formulated over $\O$ objects, where $\M \subset \O$ objects are \emph{manipulators}. The task $\T$ consists of $\S$ steps, each of which is associated with a manipulation graph $G_s$ which expresses the connectivity of objects during step $s$.
Manipulators hold a special role in interpreting $G_s$. We denote as $H_{s,m}$ the sub-graph of $G_s$ that is reachable from manipulator $m$. All objects that are not part of a manipulator subgraph $H_{s,m}$ form the \emph{world} graph $H_{s,W} = G_s / \bigcup_{m\in \M} H_{s, m}$. 
Steps $s-1$ and $s$ must differ in their manipulator sub-graphs to be separate steps, \ie $\exists m\in\M: H_{s-1,m} \not = H_{s,m}$.
For the transition between steps, a task model records a distribution over the relative states of objects whose set memberships change. If object $o\in\O$ changes membership in sub-graphs from $s-1$ to $s$, \ie $o \in H_{s,*} \wedge o \not\in H_{s-1,*}$, then the model contains $\prob{\tf{H_{s,*}}{T}{o}}$ which is the distribution over poses which $o$ will have at this transition relative to the other objects in $H_{s,*}$.
The aim of our approach is to extract such a model from a collection of demonstrations $\D = \set{d_1, \ldots, d_N}$ which have length $T_d$ and consist of trajectories of objects $\O_d \times T_d \times SE3 \times B$, where $B = \set{0, 1}$ is a boolean flag indicating the visibility of the object at the given time step. Additionally, each object $o_d$ in each demo $d$ is associated with a feature vector $f_{o_d}$. We assume that all demos in $\D$ show the same task being performed, but not necessarily with the minimal number of objects or steps. Finding the relevant objects and steps is the challenge that needs to be addressed.

\section{Task Learning Approach}
\label{sec:approach}

Our approach, presented in \cref{fig:approach}, consists of four steps:
1)~Each individual demonstration is segmented into manipulation graphs using probabilistic models.
2)~The topological changes in these graphs are used to emit three types of events which serve as the basis to learn a model across the demonstrations.
3)~Objects observed in the different demonstrations are matched using their feature vectors.
4)~The produced object match is used to associate equivalent events across demonstrations, searching for the minimum-entropy model of the task.
In the following sections, we introduce these steps in detail. Throughout, we denote hyperparameters as $\alpha$.

\subsection{Demonstration Segmentation}
\label{sec:seg_demos}

\begin{figure*}[t]
\centering

\begin{minipage}[t]{0.69\linewidth}
  \vspace{3mm}
  \centering
  \includegraphics[width=\linewidth]{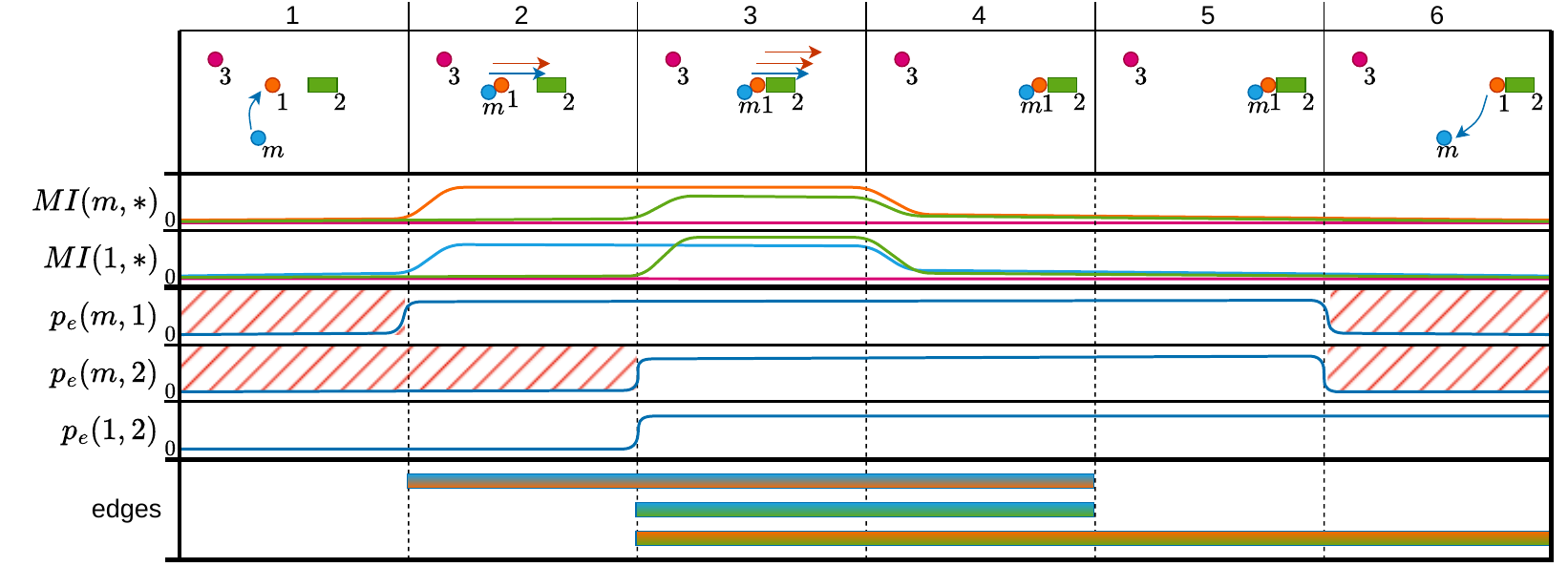}
  \captionof{figure}{ Schematic example of our segmentation approach. The manipulator $m$ moves towards object $1$ and starts pushing it towards object $2$ (frame $2$). When objects $1$ and $2$ touch, $2$ joins the pushing motion (frame $3$). The manipulation is completed in frame $4$, yet the manipulator remains in its location relative to $1, 2$ until frame $5$. Finally, the manipulator parts from the objects (frame $6$). The first three plots illustrate the $MI$ signal for these three objects and the background object $3$. Once objects are moving together, their mutual information rises. When the motion stops, the mutual information also returns to $0$. In these phases of high $MI$, the connection likelihood model $p_e(a, b)$ is formed, which scores two objects being connected based on their distances. Using this model, we identify the time steps in which objects are not being manipulated (area identified with red hatches). Using these areas to form the distribution of objects being at rest, we prune the connections $e_{a,b}$ between objects and manipulator $m$ to exclude all \emph{resting} frames at the end of the manipulation.}
  \label{fig:segmentation}
    \vspace{-5mm}
\end{minipage}
\hfill
\begin{minipage}[t]{0.29\linewidth}
  \vspace{3mm}
  \centering
  \includegraphics[width=\linewidth]{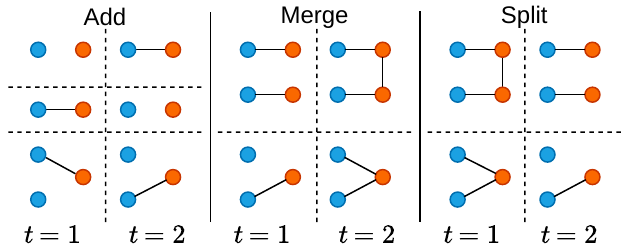}\\[0.5em]
  \includegraphics[width=\linewidth]{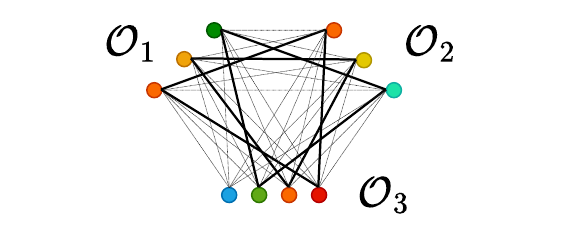}
  \captionof{figure}{\textit{Top}: Visualization of topological changes which emit events. Blue nodes represent manipulators, orange nodes represent other objects.
  \textit{Bottom}: Graphical representation of the $k$-assignment problem ($k$-ap) underlying the re-identification of objects across demonstrations. The colors represent the different features of the objects. The dotted lines indicate possible associations, bold lines show a full $3$-assignment of the given problem.}
  \label{fig:algo_problems}
    \vspace{-5mm}
\end{minipage}

\end{figure*}

Our approach begins by extracting manipulation graphs $G_s$ for a given demo $d$. As described in \cref{sec:problem}, these graphs are only generated when the nodes in the manipulator sub-graphs $H_{s,m}$ occur. Our approach first builds the moment-to-moment graphs $G_t$ and finally filters these to $G_s$. Following the approach presented in \cite{merlo2025exploiting}, we employ Mutual Information ($MI$) as a primary indicator for the correlated motion of objects. Intuitively speaking, the mutual information $MI(X, Y)$ of two distributions $X, Y$ expresses how well samples from $X$ can be used to predict the value of $Y$ and vice-versa. In our case, $X, Y$ are distributions of object poses over time. As \cite{merlo2025exploiting} explains, using $MI$ is advantageous compared to velocity-based correlations because no explicit model of the type of motion is needed. This way, linear motions can, for example, easily be connected to rotational motions.
To measure the Mutual Information of two objects $o_i,o_j$ at time $t$, we form a sliding window observation of the trajectories of $\tf{W}{T}{o_i}$ and $\tf{W}{T}{o_j}$ of length $\alpha_w$. We denote the windowed observation of $o_i$ at time $t$ as $\tf{W}{\hat{T}}{o_i} = \set{\tf{W}{T}{o_i, t-\alpha_w}, \ldots, \tf{W}{T}{o_i, t}}$. Similarly, we denote the visibility of $o_i$ at $t$ as $M_{i,t}$ and the \emph{windowed} visibility as $\hat{M}_{i,t}$.
On these windowed observations, our approach computes $MI(o_i, o_j)_t$ for step $t$, \emph{iff} both $o_i, o_j$ are visible for the entire window, \ie $\min(\hat{M}_{i,t}) = 1 \wedge \min(\hat{M}_{j,t}) = 1$.
Unlike \cite{merlo2025exploiting} we use multivariate normal distributions (MVN) as models for the sliding windows and fit them to a $6D$ representation of $SE3$ which we obtain as $f(\tf{W}{T}{o}) = \langle \tf{W}{P}{o}, \log(\tf{W}{R}{o})\rangle \in \R^6$.
We opt for MVNs over histograms as used in~\cite{merlo2025exploiting} to avoid the quantization parameter, which is context-dependent and hard to obtain for non-Euclidean observation spaces such as $SO3$.
A graphical representation of the process described hereafter is shown in \cref{fig:segmentation}.
We compute $MI(i, j)_t$ using the form for MVNs over all windowed time steps $t \in [\alpha_w .. T]$ as 
\[
    MI(i, j)_t = \frac{1}{2} \log\left(\frac{\det{\Sigma_{i,t}}\det{\Sigma_{j,t}}}{\det{\Sigma_{i,j,t}}}\right),
\]
where $\Sigma_{i,j,t}$ is the covariance of the joint pose distribution of objects $i, j$ for the window $[t-\alpha_w .. t]$. If $\hat{M}_{i,t}$ or $\hat{M}_{j,t}$ indicate the absence of observation in the window of $t$ then $MI(i, j)_t = 0$.
We use $MI(i, j)_t$ as trigger to build a connectivity model of $o_i,o_j$ by collecting data $Z_{i,j}$ as
\[
    Z_{i,j} = \fset{\tf{W}{T}{i}_t, \tf{W}{T}{j}_t}{MI(i, j)_t \geq \alpha_{MI}},
\]
while the mutual information between $o_i,o_j$ is high. Once the approach collected a sufficient number $\alpha_Z$ of observations $\cardinality{Z_{i,j}} > \alpha_Z$, it uses $Z_{i, j}$ to form a connectivity distribution
\[
    p_{e,t} = \probc{e_{i,j}}{\hat{\tf{W}{T}{i}_t}, \hat{\tf{W}{T}{j}_t}},
\]
which is used to assess the likelihood of $o_i, o_j$ being connected given the observation window ending at $t$. The connectivity model of $e_{i,j}$ is obtained as a normal distribution of the distance between $o_i, o_j$ in $Z_{i,j}$.
Naturally, objects $o_i, o_j$ can be connected and disconnected multiple times during a demonstration. Each connection creates a new edge with its own connectivity distribution. We do not allow edges between manipulators by setting \mbox{$\forall o_i \in \M \wedge o_j \in \M: p_{e,t} = 0$} in our segmentation process.
Going forward, we drop $i,j$ from $e$ and instead denote $e_{a-b}$ as an edge which is scored as connected from $t=a$ until $t=b$.
The initial edge extends from the first moment the $MI$ surpassed the trigger threshold $\alpha_{MI}$ until $p_{e,t} < \alpha_{e}$. However, the $MI$ only provides evidence that there is a connection between $i,j$, but does not mark its true beginning, \ie the grasp is completed before the object is being moved. 
To capture the inception of the connection, we propagate $e_{a-b}$ back in time to $a' < a$ while $\hat{M}_{i,t} \wedge \hat{M}_{j,t} \wedge p_{e,t} \geq \alpha_{e}$ holds. The expanded region $[a'.. b]$  represents a conservative estimate of $o_i,o_j$ being connected.
Given the connectivity, we determine at each time step whether an object $o_i$ is being manipulated by checking whether there exists a path to a manipulator. From all observed time steps where this is \emph{not} the case, we form a \emph{resting} distribution $p_{rest,i}$ which measures the observational noise in $\tf{W}{\hat{T}}{i}$.
We employ this distribution to prune $e_{a'b}$ back towards the \textit{effective} end of its manipulation \emph{iff} either of $i,j$ is a manipulator. We move $b' < b$ while $M_{i,t} \wedge M_{j,t} \wedge \min(p_{rest, i}(\tf{W}{T}{i}) \wedge p_{rest, j}(\tf{W}{T}{j}) \geq \alpha_{rest})$. 
This process yields a manipulation graph $G_t = \tuple{\O, E_t}$ for every time step $t$. %
This sequence is sparsified to a series of graphs $G_s$, which satisfy the change property between $G_{s-1}, G_s$ as described in \cref{sec:problem}.

\subsection{Event Generation}
\label{sec:event_gen}

As described in \cref{sec:problem}, graphs $G_s, G_{s+1}$ differ by the membership of objects in the subgraphs $H_{m\in\M}$ and $H_W$. However, we cannot expect these collective changes to be correlated, \ie if the task is to put a pot on a stove and cover it with a lid, pot and lid might sometimes be picked up simultaneously, sometimes this might happen asynchronously. Consequently, comparing the global change between $G_s, G_{s+1}$ is not useful. Instead, our approach generates \emph{events} for the different changes of $G_s, G_{s+1}$ and uses them to associate different demonstrations.
We define three types of events to emit:
\begin{itemize}
    \item $\eAdd(G', H_*)$: A subgraph $G'$ is added to graph $H_*$. None of the nodes in $G'$ are manipulators.
    \item $\eMerge(H_{m}, H_{n})$: Two manipulator graphs $H_{m}, H_{n}$ are connected.
    \item $\eSplit(H_m, H_n)$: The formerly connected subgraph $H_m \cup H_n$ is split into two manipulator graphs $H_m, H_n$.
\end{itemize}
\cref{fig:algo_problems} shows examples of topological changes that trigger the emission of an event. Each transition between $G_s, G_{s+1}$ can generate multiple events. Note that our approach does not permit a subgraph consisting solely of manipulators.

\subsection{Cross-Demo Object Matching}
\label{sec:obj_match}

In order to associate the generated events across demonstrations, we need to compare the different subgraphs they reference.
Let $o^n_i$ denote the $i$-th object of demonstration $n$.
Given $N$ demonstrations, we seek to re-identify as many objects as possible across all $N$ demonstrations.
We write $\phi_i = \tuple{o^1_i, \ldots, o^N_j}$ as the full association of object $i$ across all demonstrations.
We require such a match for every object in $\O^-$, which is the smallest set of observed objects across all demonstrations, and refer to this set of all matches as $\Phi = \set{\phi_1, \ldots, \phi_{\cardinality{\O^-}}}$.
To gain an intuitive understanding of this matching problem, we present a schematic in \cref{fig:algo_problems}.
Given the many ways to match objects across the different demonstrations, our approach requires a scoring function.
For this purpose, we rely on object features to $f^n_i$ and $f^m_j$, which can be generated either from visual features using DINOv2~\cite{oquab2023dinov2}, or from the objects' names using CLIP~\cite{radford2021learning}.
We employ feature similarity to score an object match $\phi_i$ as
\[
    s(\phi_i) = \sum_{n=1}^N\sum_{m=n+1}^N \log(f_n^T f_m),
\]
where $f^n, f^m$ refer to the features of the objects chosen in demo $m, n$ to be object $o_i \in \O^-$. Analogously, the score of a full match $\Phi_i$ is the sum of the scores of the individual $s_i$.
We employ the sum of logs as an analogy to the product of the similarities in order to avoid low-value associations being compensated by high-value ones.

More generally, this assignment problem is known as the $k$-assignment problem ($k$-AP), which is NP-hard for $k\geq3$~\cite{gabrovvsek2020multiple}. Despite this, there are efficient heuristics that achieve near-optimal solutions, of which we use \mbox{\emph{algorithm A}~\cite{gabrovvsek2020multiple}}.
Our sum of logs formulation is directly compatible with the standard $k$-AP form. In this graphical view, the similarity values serve as edge weights between objects from different demos (see \cref{fig:algo_problems}), with the objective being to maximize the overall chosen assignment value. While the graph seems daunting, we would like to point out that the complexity of computing $s(f_n^Tf_m)$ for all pairs is indeed lower than $O((\sum_n^N\cardinality{\O_n})^2)$.

\subsection{Extracting a Task Skeleton}
\label{sec:approach_rel_poses}

The completed object matches across demonstrations enables our approach to compare the events generated in \cref{sec:event_gen}. We refer to the series of events generated by demo $n$ as $E_n$. We assume that $E^- = \argmin_{E_n} \cardinality{E_n}$ is the best representation of our task, as it contains the fewest changes in manipulation relationships. However, it can still contain noise in object manipulations.
To filter these, we form the set of confirmed activated objects
\[\footnotesize
    \A_\T = \fset{o_i}{\alpha_\A < \frac{1}{N}\sum_n^N [\exists \eAdd(G', H_m)\in E_n: o_i \in G']},
\]
where $m \in \M$, $[\ldots]$ is $1$ if the inner expression is satisfied and $\alpha_\A$ is a decision threshold. All objects that are not in $\A$ are removed from $G'$ in all $\eAdd$ events and all $H_m$ manipulator graphs, and are returned to $H_W$ for all $\eAdd$ that reference that graph. This action can potentially change which $E_n$ is considered to be $E^-$.
Next, we collect match-candidates for all $e \in E^-$. Generally, two events match if the referenced graphs contain the exact same nodes. $\eMerge$ and $\eSplit$ are commutative, \ie $\eMerge(H_m, H_n)$ matches $\eMerge(H_n, H_m)$. A further exception is $\eAdd(G', H_W)$, for which it is sufficient for $H_{W,i} \cap H_{W,j} \not = \emptyset$. This latter case accommodates the instances in which multiple unsynchronized parallel manipulation actions are performed. Different completion orders create differing $H_W$.

Using these matching conditions, our approach selects sets of all matches for $e_i \in E^-$ in all other demonstrations. While applying $\A$ removes overall false object manipulations, it does not filter for \emph{unconfirmed object activations of a specific nature}, \ie whether a specific $\eAdd(G', H_m)$ can be found in multiple demonstrations. To remedy this fact, the sequential match finding for $e_i \in E^-$ keeps a record of the currently confirmed active objects and filters $G'$ of $\eAdd_i(G', H_w)$ for these objects. Finally, this process yields $\omega_i = \tuple{e_i, \fset{e_j \in E_2}{\psi(e_i, e_j)}, \ldots, \fset{e_j \in E_N}{\psi(e_i, e_j)}}$ for each $e_i \in E^-$ where $\psi$ is the event-matching condition.
Given the candidate collections $\omega_i, \ldots, \omega_K$, we extract a simple sequential model of the task which follows the execution of $E^-$. In this work, we focus solely on using the $\eAdd$ events, and we discuss this choice in \cref{sec:conclusion}. We identify the best-matching $e_j^n$ for an $\eAdd$ event $e_i$ by forming the power-set of candidate combinations from $\omega_i$. Each candidate $c = \tuple{e_i^1, \ldots, e_j^N}$ is scored according to the entropy of the distributions of relative poses $\tf{l}{T}{o}$ and $\tf{o}{T}{l}$ where $l\in H$ and $o\in G'$. Simply using the mean entropy would imply an equal importance of all objects, which is unlikely, especially in scenes with many objects. To account for this fact, we first compute the mean entropy $\bar{S}_i$
and its variance of all object distributions $\tf{l}{T}{o}$ and $\tf{o}{T}{l}$ for all combinations in $\omega_i$. This allows us to score the probability of the entropy of each distribution $\prob{\tf{l}{T}{o}}$ given the candidate set, where we clamp deviations from $\bar{S}_i$ to $\leq 0$. We choose the candidate $c_i^*$ with the lowest average entropy probability as the best candidate for event $e_i$.
Our sequential task model $\T$ consists of a series of relative pose distributions for objects, switching between $G'_t$ and $H_t$. We reduce the number of objects $l$ considered in each step according to the entropy of $\tf{l}{T}{o}$. All our sequential models are simple MVNs capturing $\probdist{t}{\tf{W}{T}{o}} = \prod_{l \in H} \probc{t}{\tf{l}{T}{o}}\probc{t}{\tf{o}{T}{l}}$ for each %
step $s$.

\subsection{Inference}
\label{sec:encoding_choice}

\begin{figure*}
\begin{minipage}[t]{0.69\linewidth}
    \vspace{1mm}
    \centering
    \includegraphics[width=0.95\linewidth]{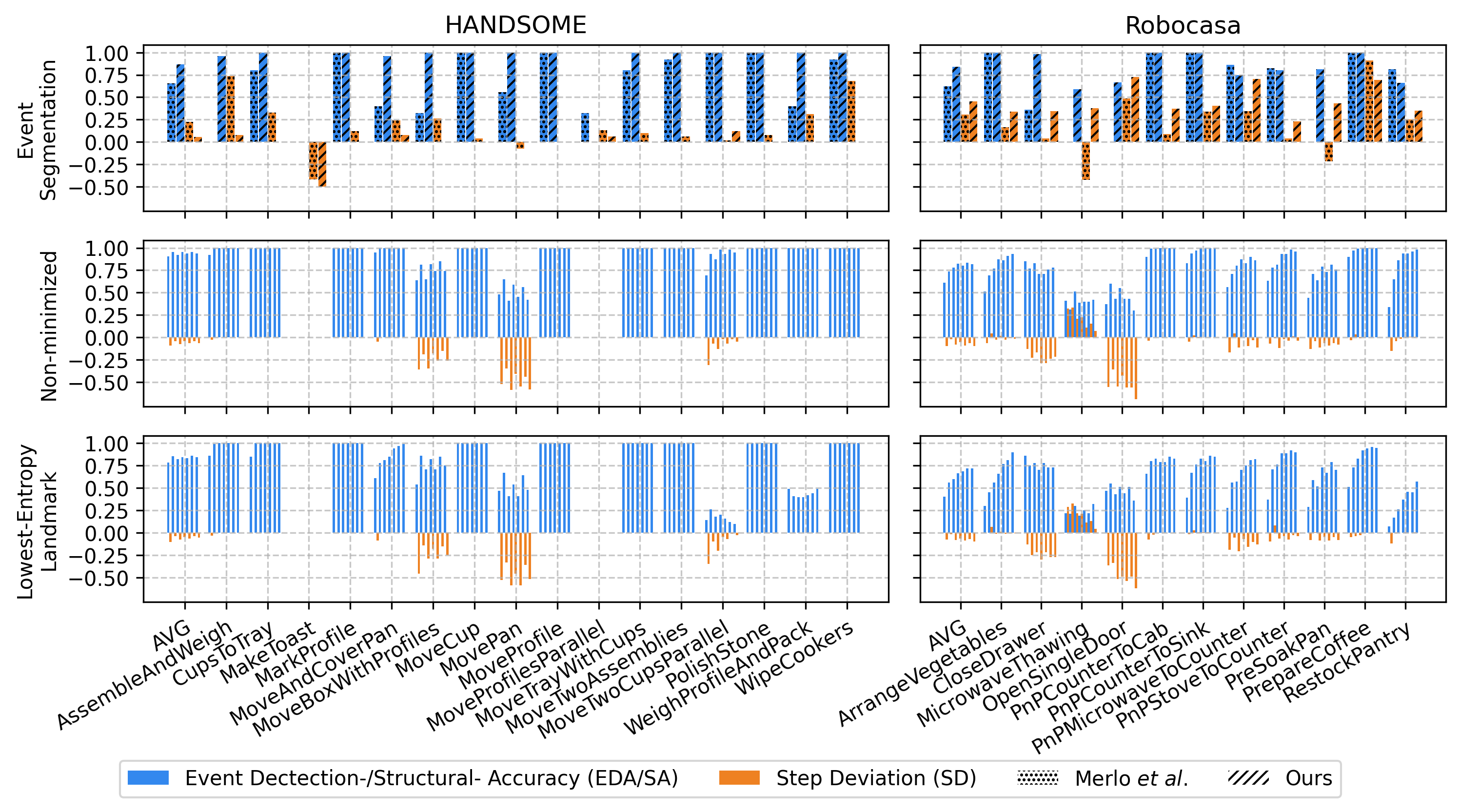}
    \vspace{-2mm}
    \caption{\textit{Top Row}: Results of segmentation and event generation. We report the normalized Event Detection Accuracy (EDA) and over-/under segmentation (SA) where $0$ is ideal. Our approach exhibits an average success rate of $85\%$, while showing a tendency to over-segment by $20\%$. Over-segmentation is more pronounced on Robocasa data, which includes many failed or accidental manipulations.
    \\\textit{Bottom Rows}: Results of extracting task models using our proposed approach. We sample a random $8$ demonstrations for each task and then fit a model from these, starting with $2$ and incrementally increasing to the full set. We sample $100$ of these sets per task and report the averaged metrics. \emph{Middle}: Success of our approach at extracting the correct model steps. We observe that adding more demonstrations improves the accuracy of the fitted model, with saturation occurring quickly. \emph{Bottom}: For all steps $Add(*, H_W)$ the reference group $H_W$ is reduced to the minimal entropy object. This lowers the model accuracy. In HANDSOME, the effect is limited to a few tasks, but dramatic. In Robocasa, the effect is more spread out and mostly delays the saturation of yield through additional demonstrations.}
    \label{fig:all_results}
    \vspace{-3mm}
\end{minipage}
\hfill
\begin{minipage}[t]{0.29\linewidth}
    \vspace{1mm}
    \centering
    \includegraphics[width=\linewidth]{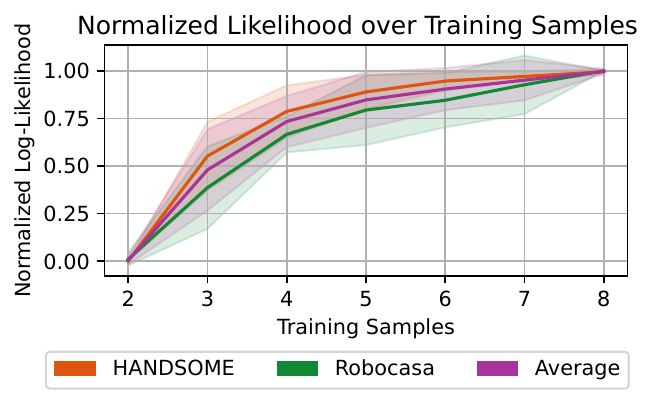}\\[0.5em]
    \includegraphics[width=\linewidth]{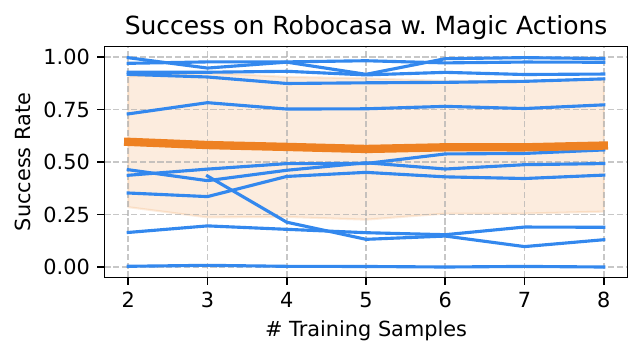}
    \captionof{figure}{\textit{Top}: We compare the likelihood of fitted models against a test set of demonstrations, given different numbers of training samples. We normalize by the overall change in likelihood within one experiment. We see additional training samples quickly yielding diminishing improvements in the likelihood.
    \textit{Bottom}: Evaluation of structurally correct, minimized models on Robocasa with magic actions. We note that performance is constant across training samples but differs strongly between tasks (blue lines).}
    \label{fig:task_likelihood}
    \vspace{-3mm}
\end{minipage}
\end{figure*}

For inference, we identify the necessary objects from a set of candidates using the object features stored with our model.
With the identified objects, we generate a plan to execute the task. Given that all our step models are a combination of unimodal Gaussian distributions, we can employ optimization to maximize the likelihood of an object pose for each step. Using a differentiable kinematics framework~\cite{roefer2022kineverse} we include kinematic feasibility through expressing the likelihood of a configuration as a function of a robots joint angles. Thus, we solve for the robot joint states to minimize the negative log-likelihood of the object's pose.

Our models do not include any information about the type of manipulation associated with each event, \ie whether an object is being grasped, pushed, or scooped.
For the current inference, we follow a simple rule for interpreting $\eAdd(G', H)$ events: If $H$ is a manipulator graph and $\cardinality{H} = 1$, \ie the manipulator is free, we close the robot's gripper after it has reached the desired position, interpreting the beginning of any manipulation to be a grasp. If $H$ is the world graph and $\cardinality{H_m} = 1$ after the event, we open the gripper at the end of the robot's motion. While an object is being manipulated, we assume a fixed relative transform to the manipulator.

\section{Experimental Evaluation}

We evaluate our approach on HANDSOME~\cite{merlo2025exploiting} and Robocasa~\cite{robocasa2024} to first generate demonstration segmentations and, secondly, full task models. We then gauge the utility of these models by deploying them in a simulated environment and using them for inference on a physical robotic system.
We keep our hyperparameters constant across our evaluation as $\alpha_w=8, \alpha_Z=3, \alpha_{MI}=0.18, \alpha_e=0.1, \alpha_\A=0.5$. We swept the most significant $\alpha_w \in [5..15]$ and $\alpha_{MI} \in [0.1..0.4]$ and did not find a significant difference in results. Due to significant noise in the HANDSOME dataset, even after third-order smoothing, we only use positional observations for the initial segmentation.

\subsection{Segmenting Demonstrations}
\label{sec:exp_segmentation}

The first part of our approach, \cref{sec:seg_demos}, segments demonstrations as per the problem definition in \cref{sec:problem}. To the best of our knowledge, no other method with this exact aim exists. In order to contextualize the results of our method, we compare with~\cite{merlo2025exploiting}, which we re-implement as per the detailed specification given in the article. We use the dynamic manipulator-object and object-object relationships which \cite{merlo2025exploiting} identifies to build $G_t$ and generate events from these as per \cref{sec:event_gen}.
Datasets with 3D trajectory observations for this problem domain are in short supply. We use the HANDSOME dataset originally introduced with~\cite{merlo2025exploiting} and additionally obtain trajectory observations from the Robocasa~\cite{robocasa2024}. This selection yields $950$ demonstrations across $27$ tasks.
Neither data source provides labels for interaction-phases, thus we create labels of events that are expected to be extracted for the given demonstrations, and measure the methods' performances in generating these in two ways: \emph{Event Detection Success} (EDS) is a binary metric which is $1$ if the approach generates all expected events in the correct order. \emph{Segmentation Divergence} (SD) measures the delta of the generated events against the labeled count. The ideal value of this metric is $0$, with positive values indicating over-segmentation and negative values indicating under-segmentation. We report this number normalized by the labeled event-count to make it comparable across demonstrations of different lengths.

\textbf{Results}: \cref{fig:all_results} shows a visual overview of the average results of either approach for demonstration segmentation. Overall, both methods successfully generate graph sequences that exhibit the labeled events. On average, \cite{merlo2025exploiting} does so in $65\%$ of cases, while our approach does so in $85\%$ of cases.
The lower segmentation success of the baseline is largely due to two factors: it relies on distance thresholds to determine when two objects are in interaction, and it considers only objects as manipulated when the MI between the object and a manipulator is high. The latter indicates the end of a manipulation when an object stops moving, independent of whether the manipulation of that object is complete. Setting the distance threshold is difficult for all types of object-object and object-manipulator pairings. We have done our best to set the thresholds for the data at hand, but especially the oblong objects, \ie drawers and doors, in the Robocasa tasks pose a challenge for this heuristic.
Our proposed approach does not face these challenges due to the edge-likelihood models that it estimates from the data. While this works very well in most cases, it can also lead to failures: In the \emph{MoveProfilesParallel} task, our approach's lack of distance constraint forms a connection between two objects that are being moved perfectly parallel by two independent hands without any actual physical connection between them.
Both approaches fail completely on \emph{MakeToast}, a task that shows one hand clashing a cup against another. The joint motion caused by this collision is too short and small to be recognized by either method.
In over- and under-segmentation, we observe large variance across methods depending on the dataset. On HANDSOME, the baseline yields minor over-segmentation on average, largely due to two large outlier tasks. On Robocasa, our method demonstrates a large over-segmentation. Upon investigation, we find these are largely due to accidental manipulations, such as bumping into a cabinet door, which our method segments despite their brevity. For our baseline, we use the recommended window size of $40$ observations to help suppress this noise during execution.

\begin{figure}
    \centering
    \vspace{1mm}
    \includegraphics[width=\linewidth]{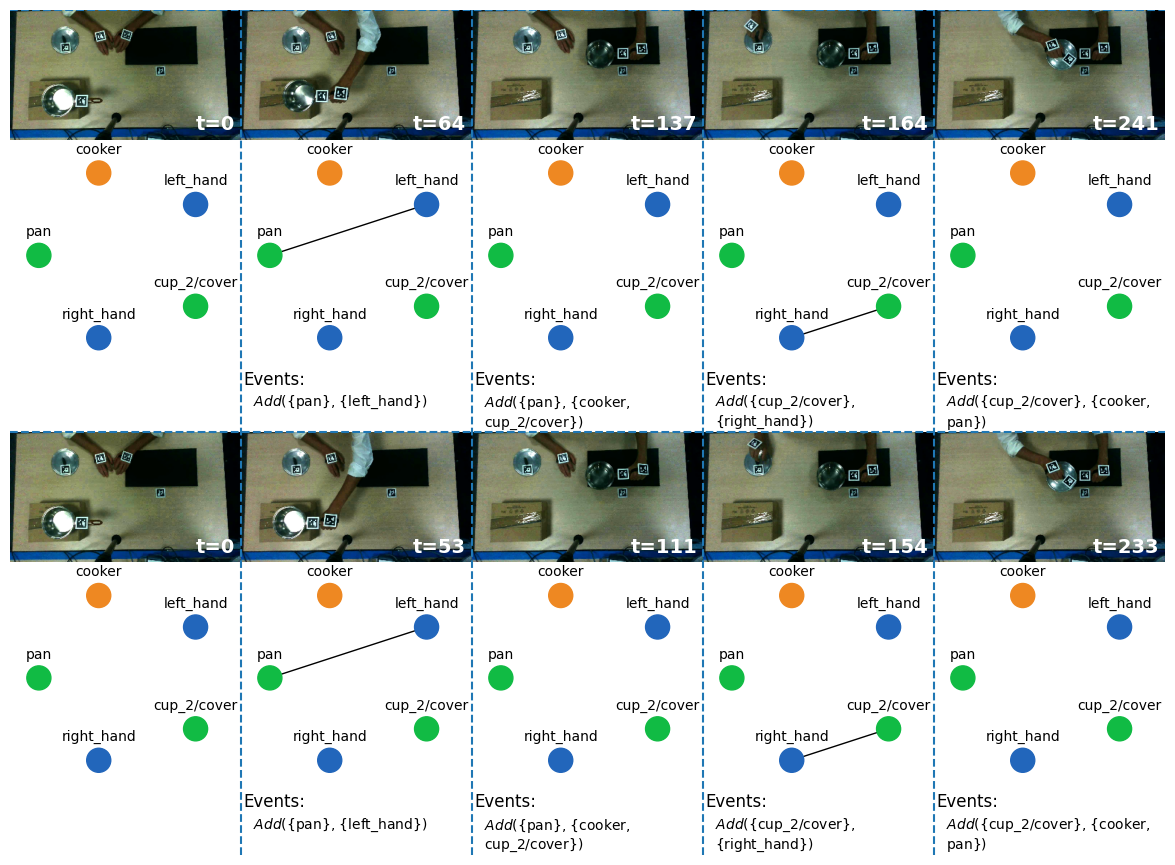}
    \caption{Example timing difference between \cite{merlo2025exploiting} (\emph{top}) our approach (\emph{bottom}). By back-propagating the edges before the observed MI spike according to their likelihood models, our approach identifies the picking up of manipulated objects (green) earlier. Similarly, the ends of the manipulation actions are also segmented earlier by exploiting the fitted \emph{resting distributions}.}
    \label{fig:qualitative_comparison}
    \vspace{-6mm}
\end{figure}

\subsection{Task Model Fitting}
\label{sec:eval_task_fitting}

To evaluate the utility of our task model fitting, we create a \emph{structurally} optimal task model for each task in our dataset. Each model contains the minimal number of manipulation steps and objects relevant to those steps. We test whether the approach can produce these steps and how many additional steps it produces. The first of these criteria we call the \emph{Structural Accuracy} (SA), for the second we reuse the \emph{Step Deviation} (SD) from \cref{sec:exp_segmentation}. Ideally, we would see the SA to approach $1$ as we increase the number of demonstrations and SD to approach $0$ at the same time. To understand the utility of the demonstrations for model minimization, we make the SA criterion harder, such that all reference groups $H_*$ have to be an exact match instead of a simple subset in the case of $H_W$. For the tasks at hand, a single reference object should be sufficient for these goal distributions, thus we simply reduce $H_W$ to the lowest-entropy object.

\textbf{Results}: We compare both the non-minimized models and the final minimized models in \cref{fig:all_results}.
In both conditions, we observe that the HANDSOME dataset contains very little noise during execution, so adding additional demonstrations does little to improve structural quality or minimize segmentation error. In the Robocasa demonstrations, the effect is present and saturates after $5$ demonstrations in the case without minimized models, and after $7$ demonstrations with the minimization heuristic. Structural soundness says nothing about the utility of the distributions fitted under our model, \ie if they can also place objects reasonably. Instead of a per-instance prediction delta, which assumes that only a single inference is valid for any given demonstration, we study the change in likelihood of our models given test data. For any given training set of up to $8$ demonstrations, we use the remaining ones as a validation set. We match the extracted model against the events of the remaining demonstrations and measure the model's likelihood given the observations at the time the event is emitted. As likelihoods have no nominal value, we compare the change in likelihood as the number of training samples rises, which we present in \cref{fig:task_likelihood}. Moreover, we observe a saturating effect, with most of the improvement occurring across the three additional samples. Lastly, on Robocasa we evaluate the inference by deploying our structurally sound models inside the simulation and realizing inferences with magic actions, \ie direct placements without robotic execution. We observe that the nominal performance varies strongly across tasks, with some performed almost perfectly and others failing severely. The average success is $57\%$ with pick and place tasks, performing the best out of the range. Our inference scheme does not consider physical plausibility, which causes object intersections resolved by scenes exploding. In the case of articulated objects, our approach struggles to extract a suitable reference object, leading to nonsensical target poses.

\textbf{Failures}: From \cref{fig:all_results}, it is apparent that performance is bimodal: in most cases, our approach is very successful at extracting task models, whereas in others it seems to work barely at all. Even more surprising: Adding more demonstrations seems to exacerbate these problems. We identify three main causes for these failures:
1) \emph{Simultaneous Bi-Manual Manipulations}: While most of the HANDSOME tasks can be extracted perfectly, \texttt{MoveBoxWithProfiles} and \texttt{MovePan} seem to pose a great challenge. \cref{fig:all_results} shows that the segmentation is not the culprit. The poor performance we observe stems from the coordinated nature of the task. In both tasks, there is a central object being grasped by both hands, and it is then moved to a different location. Each grasp creates an $\eAdd(G', H_A)$ followed by $\eMerge(G' \cup H_A, H_B)$. Since grasping and letting go happen almost simultaneously for both hands, it is very likely to encounter both sequences $A,B$ and $B,A$. Given our matching criteria for events, these different orders cannot be associated, which leads to the low structural accuracy in both tasks.
2) \emph{Reliable Imperfect Manipulation}: This error is similar to the sequencing problem in the simultaneous bi-manual manipulation scenarios. In \texttt{MicrowaveThawing}, the robot has to open a microwave, take a food item from a plate, place it on a different tray in the microwave, and close the door. In this process, the robot often moves either of these trays, but not reliably. This leads to either the pick-up or the placement of the food item to not be extracted as its own step, but in combination with the manipulation of either tray. This violates our labeled ideal.
3) \emph{Variance as Prerequisite}: We observe a significant drop in model accuracy when we perform the model minimization step. This hints at a lack of randomization in the training data. The tasks \texttt{WeighProfilesAndPack} and \texttt{MoveTwoCupsParallel} experience the largest overall drop. In both cases, objects are moved to or between two possible goal locations. Yet, the relative pose between these locations only varies slightly, so that the variances with respect to either are almost identical. In these cases, it is impossible to understand which of the two is the correct point of reference. 

\subsection{Real Robot Experiments}

To understand the utility of our approach for robotic manipulation, we embed our fitted models into a larger robotic execution system and use it to reproduce three HANDSOME tasks. As our approach operates directly on object frames, we only need to define a mapping once between HANDSOME objects and our real objects, allowing us to perform zero-shot transfer of task models. For our execution, we use a PR2 as our robotic platform (see \cref{fig:approach}), which regrettably only has one working arm. This limits us to tasks that can be executed sequentially with a single manipulator. As the distributions our approach captured for grasping objects are not directly transferable to our robot embodiment, we replace them with a different set of grasping distributions we capture in addition. We use a QP-based controller that generates joint velocities to maximize the probability of the current object pose. The controller also limits high end-effector speeds and encodes a simple lifting behavior for placing, and an approach behavior for grasping. The next stage of the task is always triggered when the probability maximization anneals. As in HANDSOME, we track our objects using Aruco markers, but are not limited to them. Marker-less options like FoundationPose~\cite{foundationposewen2024} could also be used.
We share an impression of the executions in the video attachment. 
We find that the model inference is stable, and that one can vary the object locations in the robot's workspace. The object-centric nature also enables the robot to adapt to changes in object location mid task execution. 

\section{Conclusion}
\label{sec:conclusion}

In this work, we presented an approach for obtaining sparse task-skeletons from a handful of demonstrations. Our approach competently segments demonstrations into a sparse series of manipulation graphs and generates events from topological changes within them. These can be matched across demonstrations, using pre-trained object features to minimize over-segmentation in tasks and extract skeletons of steps with pose distributions for these. Through our evaluation, we demonstrate applicability to real-world robotics.
Nevertheless, there remains ample room for improvement. As the detailed analysis in \cref{sec:eval_task_fitting} reveals, ignoring $\eMerge$ and $\eSplit$ events during event matching poses a significant obstacle to extracting skeletons for coordinated bimanual tasks.
From a practical perspective, the remaining dependence on some hyperparameter tuning hinders hassle-free deployment. Most of our $5$ hyperparameters are not sensitive, but setting $\alpha_{MI}$ is dependent on the width of the distributions fitted inside the observation windows, which correlates with the scale of the scene and objects' speeds. Being able to infer this value from the data itself would be a leap for methods like ours or~\cite{merlo2025exploiting}.
For future work, our approach also collects a lot of information, which we do not exploit to the fullest at the moment. One example of this is the overall likelihood within a manipulator sub-graph, and also the tightness of these distributions. Interpreting these could yield true kinematic trees that more closely model the (dis-)connectedness of objects. Studying tool-use cases, such as a drill bit being placed in the socket of a power drill or a pancake being flipped with a spatula, would be interesting. %

\bibliographystyle{IEEEtran}
\bibliography{sources}  %

\end{document}